\documentclass[10pt,twocolumn,letterpaper]{article}

\usepackage{cvpr}              % To produce the CAMERA-READY version
\usepackage{graphicx}
\usepackage{amsmath}
\usepackage{amssymb}
\usepackage{booktabs}
\usepackage{changepage} 
\usepackage{tabularx}
\usepackage{float}

\usepackage[pagebackref,breaklinks,colorlinks]{hyperref}

\usepackage[capitalize]{cleveref}
\crefname{section}{Sec.}{Secs.}
\Crefname{section}{Section}{Sections}
\Crefname{table}{Table}{Tables}
\crefname{table}{Tab.}{Tabs.}

\def\cvprPaperID{8422} % *** Enter the CVPR Paper ID here
\def\confName{CVPR}
\def\confYear{2022}

\begin{document}

\title{DeePaste -  Inpainting for Pasting}

\author{Levi Kassel\\
The Hebrew University of Jerusalem\\
Jerusalem, Israel \\
{\tt\small levi.kassel@mail.huji.ac.il}
\and
Michael Werman\\
The Hebrew University of Jerusalem\\
Jerusalem, Israel \\
{\tt\small michael.werman@mail.huji.ac.il}
}
\maketitle

%%%%%%%%% ABSTRACT
\begin{abstract}
One of the  challenges of supervised learning training is the need to procure  an substantial amount  of tagged data. A well-known method of solving this problem is to use synthetic data in a  copy-paste  fashion, so that we  cut  objects  and  paste  them onto relevant  backgrounds. Pasting the objects naively results in artifacts that cause models to give poor results on real data. We present a new method for cleanly  pasting  objects on different backgrounds so that  the dataset created gives competitive performance on real data. The main emphasis is on the treatment of the border of the pasted object using inpainting. We show state-of-the-art results  both on instance detection and foreground segmentation.
\end{abstract}

%%%%%%%%% BODY TEXT
\begin{figure*}
  \includegraphics[width=\textwidth,height=4cm]{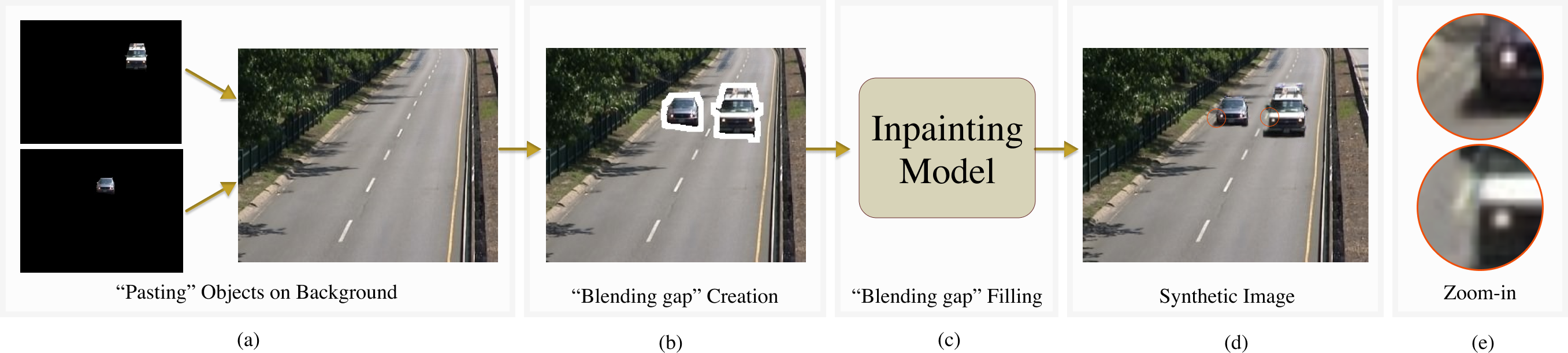}
  \caption{This is the main pipeline of our system. First we  paste  objects on the background image (a). Second, we create the  blending gap  mask (b). We use the trained inpainting model to fill the  blending gap  pixels (c). In (d) you can see the final image. The pixels are filled with the  right  context (e). 
  }
  \label{fig:main process}%
\end{figure*}

\begin{figure}
  \includegraphics[height=13cm]{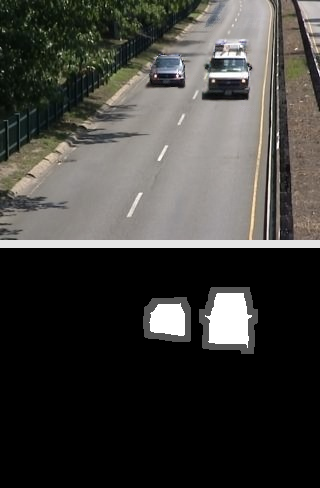}
  \caption{The  pasting  in the top image is over the white masks. The  inpainting is  on the gray area.}
  \label{fig:foreground segmantation evaluation}%
\end{figure}
\section{Introduction}
\label{sec:intro}

In recent years, with the dramatic development of detection and segmentation
in computer vision with the help of AI, the demand has increased to put these capabilities into existing systems such as mobile apps, autonomous vehicles, robots, etc.
One of the biggest problems with state of the art systems is the  amount of tagged data needed for each task or scene to train them.

For each new task, as well as for any change in the environment, such as a change in the objects you wish to detect or detecting objects in a different scene, one is required to create thousands or more tagged images based on the same task or scene. Since data tagging is very labor intensive, deploying those tasks is not carried out in most cases.

In recent years more and more researchers have been able to overcome this challenge by creating synthetic data from  synthetically rendered scenes and objects\cite{hinterstoisser2018pre}\cite{rozantsev2015rendering}\cite{hinterstoisser2019annotation}\cite{rajpura2017object}, so that they render objects and scenes for training detection and segmentation systems.
Although these systems bring promising results, they require a high level of graphics know how and a lot of effort to make objects and scenes look  real .

Moreover, models that train on such data find it difficult to give satisfactory results on real data because of the change in the statistics of the image\cite{chen2016synthesizing}\cite{peng2015learning}.
To overcome this difficulty and to still  generate tagged data faster, more and more researchers are working on composing  real  images \cite{dwibedi2017cut}\cite{wu2018busternet}\cite{ghiasi2021simple}. The general idea is to  cut  real objects and  paste  them on real scenes, thus getting free tagged images that fit the environment in which we want to work.

However, if we use this paradigm naively - we will get poor results on real data. This is because existing systems are more sensitive to local region based features than to global scene layout.
The implication is that when we paste an object on an image we are creating
subtle pixel artifacts on the background images. This phenomenon prevents the system from generalizing to real images.
Dwibedi et al. \cite{dwibedi2017cut} create a framework that combines all kinds of simple blending and blurring types on the objects and achieves surprisingly impressive results.

In this article, we present a new method of blending to overcome the difficulty of local-level realism.
We use the power of inpainting  trained on relevant   scenes and thus succeed in filling in the gaps in  required areas - in our case - the area between the object and the background, the {\it blending gap}.

Because we know  the environment  in which we want to work,  we  train an  inpainting model that  knows  how to fill the  blending gap  so that the detection and segmentation systems can be generalized to real data.

In recent years there have also been  works on improving the  blending  and improving the  harmonization  of objects that are  pasted  on a background image such as Deep Harmonization\cite{tsai2017deep}\cite{cong2020dovenet} or Deep Blending \cite{wu2019gp}\cite{zhang2020deep}. These systems were not compared because the purpose of these systems is graphical, and perform the learning on each image and object individually\cite{zhang2020deep} - which makes the purpose of creating a large dataset  non-feasible. What we do is different. The images we produce are not graphically perfect - but are good enough to  cheat   the learning systems so that they can generalize well on real data.

In this article, we evaluate this method and its effectiveness and show that it achieves state of the art results on instance detection and foreground segmentation.

\begin{figure*}
  \includegraphics[width=\textwidth,height=4cm]{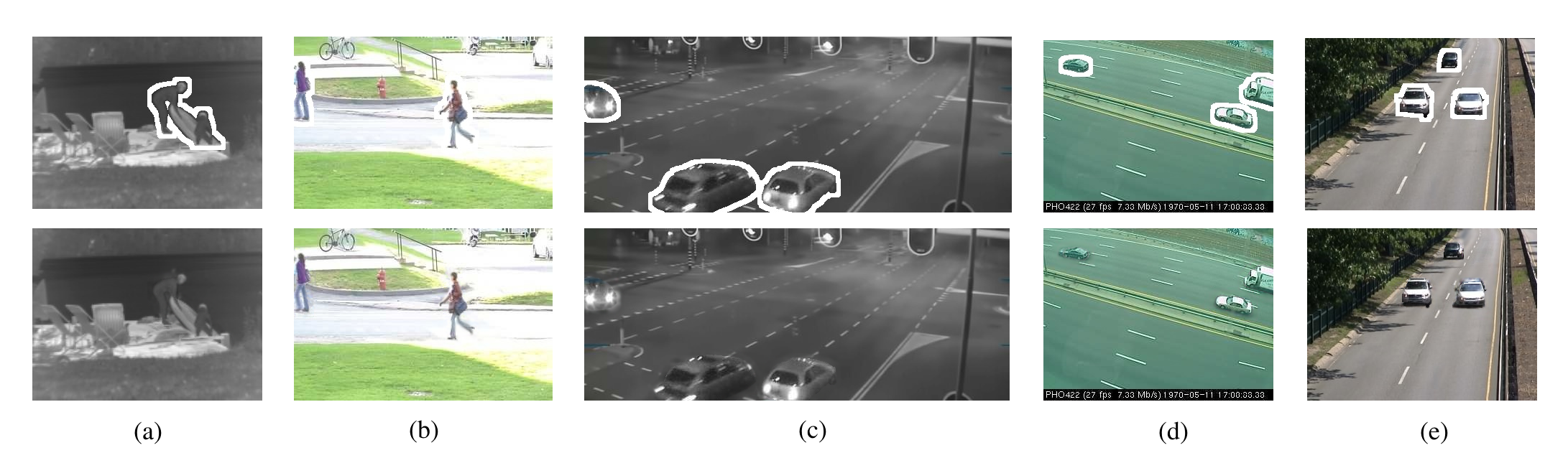}
  \caption{ Sample pairs of  blending gap  filling  from the CDnet 2014 dataset\cite{wang2014cdnet} in thermal (a), baseline (b,e), night videos (c), and low frame-rate (d) scenes.}
  \label{fig:foreground segmentation}%
\end{figure*}
%-------------------------------------------------------------------------
\maketitle
\section{Related Work}
\label{sec:intro}

Instance detection is  where we try to locate a particular object  within a target image. 
Early approaches, such as \cite{collet2011moped}  tried to solve this task by using local features of the desired object image such as SIFT \cite{lowe2004distinctive}, FAST \cite{fas}, or SURF \cite{bay2006surf}, and try to match them with local features  extracted from the target image. These methods did not work well when the objects we wanted to detect were partially occluded or did not have enough  features \cite{hinterstoisser2011gradient}\cite{hsiao2010making}.

Another well-known computer vision task is foreground segmentation (also called background subtraction),  determining which pixels  belong to the background and which  to the foreground, mainly from videos. Early approaches, such as \cite{stauffer1999adaptive}\cite{st2014subsense}\cite{van2012background}\cite{barnich2010vibe}, were mainly  based on analyzing the displacement of certain parts of the image and  trying to decide if it is a displacement of a foreground object or a natural change that occurs in the background such as shading, change in lighting, or natural displacement of dynamic objects belonging to the background.

Modern methods in these two areas are mainly deep learning systems based on convolution network architectures so that the extraction of the features is done more  semantically  even in difficult cases \cite{ren2015faster}\cite{redmon2018yolov3} \cite{lim2020learning}\cite{wang2017interactive}. Most of the best performing systems are supervised - which means that they learn directly from tagged images and are trained end to end. 

In recent years, with the availability of powerful hardware, these methods are more suitable for real-time applications and thus the demand is increasing to use these capabilities in more and more areas like robots, navigation, and surveillance \cite{watson2017real}\cite{minaee2021image}.

These methods require large amounts of annotated data. There is a bottleneck as each environment and scene requires a collection of annotated data  which significantly slows the deployment of these systems.

One method that solves this problem is to create a synthetic dataset. One setting where synthetic data can be generated is by rendering the environment  and thus automatically creating a dataset tailored to the specific problem we want to solve\cite{hinterstoisser2018pre}\cite{rozantsev2015rendering}. A big advantage of this method is that the space of possibilities is large - you can create any environment you want and you can change the objects and background simply as needed \cite{hinterstoisser2019annotation}\cite{rajpura2017object}. Unfortunately, when trying to generalize those systems from synthetic data to real would images - we encounter poor performance due to the difference in the statistics between the images space\cite{chen2016synthesizing}\cite{peng2015learning}.

Another method to produce synthetic data is by composing real images. In general, the method works by taking existing objects, which you want to detect or to segment,  cut  them, with a corresponding image mask, select a background image, and  paste  the object in a certain position in the background image. There are certain algorithms where you can automatically choose where to paste an object  to maintain the  realism  of the created image \cite{tripathi2019learning}\cite{kassel2020using}. This process is very scalable and  you can easily create a variety of data for the required purpose.

Dwibedi et al.\cite{dwibedi2017cut} took this process and examined it on the task of instance segmentation. In their paper, they notice that modern deep learning based systems care more about local region-based features for detection than the global scene layout. This is why it is very important to pay attention to how the  pasting  process takes place. When we naively place objects in scenes 
subtle pixel artifacts are introduced into the images. As these minor imperfections in the pixel space feed forward deeper into the layers of a ConvNet \cite{lecun1989backpropagation}, they lead to noticeably different features and the training algorithm focuses on these discrepancies to detect objects, often ignoring to model their complex visual appearance.
To overcome this problem they used blending methods such as Gaussian Blurring and Poisson editing \cite{perez2003poisson} and found that the use of a blending method in the  pasting  process can alter the detection performance dramatically. In addition, they found that if they synthesize the  same scene with the
same object placement only varying the type of blending used  makes the training algorithm
invariant to these blending factors and improves performance by a big margin.

Kassel et al\cite{kassel2020using} used this framework in  foreground segmentation.
They used a  weak foreground segmenter to extract objects from the training images and insert
them in their original position into a background image. It
is especially pertinent to static cameras as the objects found
are automatically in the  right  location, being of the  right 
size, color, and shape, and in the  right  lighting conditions.

Another task in computer vision that is related to image synthesis is image inpainting.
The main idea of image inpainting is to fill missing pixels  in the image.
The main difficulty in this task is to synthesize visually realistic and semantically plausible pixels for the missing regions that are coherent with existing ones.

Early work on this subject \cite{barnes2009patchmatch}\cite{hays2007scene} attempted to solve this task by searching for similar areas in the image  and attempting to  paste  patches into the missing holes while maintaining global consistency.
These works work well especially in images that are characterized by stationary textures but are limited to non-stationary data such as natural images\cite{simakov2008summarizing}.
In recent years, deep learning has also entered this field and GAN-based\cite{goodfellow2014generative} approaches have yielded promising results\cite{iizuka2017globally} \cite{li2017generative}.
Yang et al. \cite{yang2017high} and \cite{yu2018generative} devised feature shift and contextual attention operations, respectively, to allow the model to borrow feature patches from distant areas of the image. Another work by  Yu et al. \cite{yu2019free} tries to handle irregular holes by  filling  using gated convolutions. 
%-------------------------------------------------------------------------

\section{Approach Overview}
\label{sec:intro}
As in  {\it Cut, Paste, and Learn} \cite{dwibedi2017cut}, we focus only on the process of  pasting  objects on the background images, as part of the creation of synthetic data.
From their paper, it can be seen  that there is a high correlation between  the  pasting  process  and how the system focuses on the appearance of the  pasted  object and so manages to generalize well on real data.
As mentioned, if you paste the object somehow on the background image, there are pixel artifacts in the gap between the object and the background image which greatly affects how the system  treats real data.
In other words, the object is fine, the background is fine, but the border between them causes problems.
It can be  seen that if we perform even just Gaussian smoothing at the edges of the object - the generalization significantly increases\cite{dwibedi2017cut}.
In this paper, we  used this insight, and the power of inpainting, to make this connection as  natural  as possible in terms of convolution networks.

Here we  list the stages we did to enhance the  pasting  process.

\begin{itemize}
\item{\bf Train an inpainting model}
We train a model to learn how to  paste  objects on background images in a  natural  way. We train an inpainting model that learns the statistics of the scene and can fill designated  holes   naturally . We tune our model in filling the  blending gap  as well as in filling irregular  holes  in this environment.
\item{ \bf Paste objects on background}
When pasting  the object on the background image we automatically create  a  blending gap  mask that separates the object from the background image, Figure \ref{fig:main process} (b).
\item{\bf Fill the  blending gap }
We use the inpainting model trained in Stage 1 to fill in the  blending gap, Figures \ref{fig:main process} (c,d).
\end{itemize}

%-------------------------------------------------------------------------
\section{Approach Details}
\label{sec:intro}
We now present our approach in detail.
\subsection{Train an inpainting model}
To fill the  blending gap  correctly we need to use a model that can fill this area so that it will be as  natural  as possible. this model needs to have the capability to harmoniously  fill many \emph{irregular}  holes  in the image. For that we chose to train the DeepFill-V2 inpainting model\cite{yu2019free}, which is designed especially to deal with filling \emph{irregular} structures. This model uses a simple algorithm to automatically generate random free-form masks on-the-fly during training. So that the model will succeed to fill the  blending gap  correctly we train it on the  background images. In addition to enhancing the inpainting model's capabilities specifically on the  pasting  work, we add to the training set some fixed mask around real objects, with a weak segmenter, such that filling the  blending gap  will be fine-tuned.

\begin{table*}[htb]
\scriptsize
\centering % used for centering table
\begin{tabular}{|l|c|c|c|c|c|c|c|c|c|c|c|c|} % centered columns (4 columns)
\hline\hline %inserts double horizontal lines
\textbf{Category} & \textbf{Coca} & \textbf{Coffee } & \textbf{Honey}& \textbf{Hunt's} & \textbf{Mahatama} & \textbf{Nature} & \textbf{Nature} & \textbf{Palmolive } & \textbf{Pop}& \textbf{Pringles} & \textbf{Red} & \textbf{mAP} \\

\textbf{} & \textbf{ Cola} & \textbf{ Mate} & \textbf{ Bunches}& \textbf{Sauce} & \textbf{Rice} & \textbf{V1} & \textbf{V2} & \textbf{Orange} & \textbf{Secret}& \textbf{BBQ} & \textbf{Bull} & \textbf{} \\
%heading
\hline\hline % inserts single horizontal line
  No blending                          & 64.3 & 87.4 & 83.5 & 57.9 & 61.4 & 92.3  & 79.4 & 59.2 & 54.8 & 44.4 & 32.2 & 65.2 \\
  Gaussian Blurring                    & 65.4 & 86.4 & 80.4 & 69.3 & 64.3 & 90.3  & 83.4 & 59.3 & 59.4 & 66.4 & 42.5 & 69.8   \\
  Poission                              & 63.6 & 83.2 & 67.3 & 55.3 & 29.2 & 85.5  & 66.4 & 60.3 & 70.8 & 50.3 & 19.0 & 59.2   \\
  Inpaint(Ours)                        & 70.3 & 90.4 & 84.0 & \bf{73.7} & 62.4 & \bf{94.3}  & 83.5 & 73.4 & 65.4 & 68.3 & 44.5 & 73.7   \\
  \hline % inserts single horizontal line
   All Blend+Same Image(No Inpaint)   & 77.3 & 91.0 & 79.3 & 69.3 & 67.2 & 92.9  & 81.2 & 65.2 & 77.1 & \bf{71.8} & 40.2 & 73.9   \\
  All Blend+Same Image(With Inpaint) & \bf{80.4} & \bf{93.0} & \bf{84.6} & 73.3 & \bf{72.9} & 94.1  & \bf{88.4} & \bf{75.4} & \bf{78.9} & 66.8 & \bf{47.9} & \bf{77.8}   \\
  
\hline %inserts single line
\end{tabular}

\caption{ Evaluation results on the GMU Dataset \cite{georgakis2016multiview} from models that were trained with different  pasting  techniques.}
\label{table:instance-detection}% is used to refer this table in the text
\end{table*}

\begin{figure*}
  \includegraphics[width=\textwidth,height=24cm]{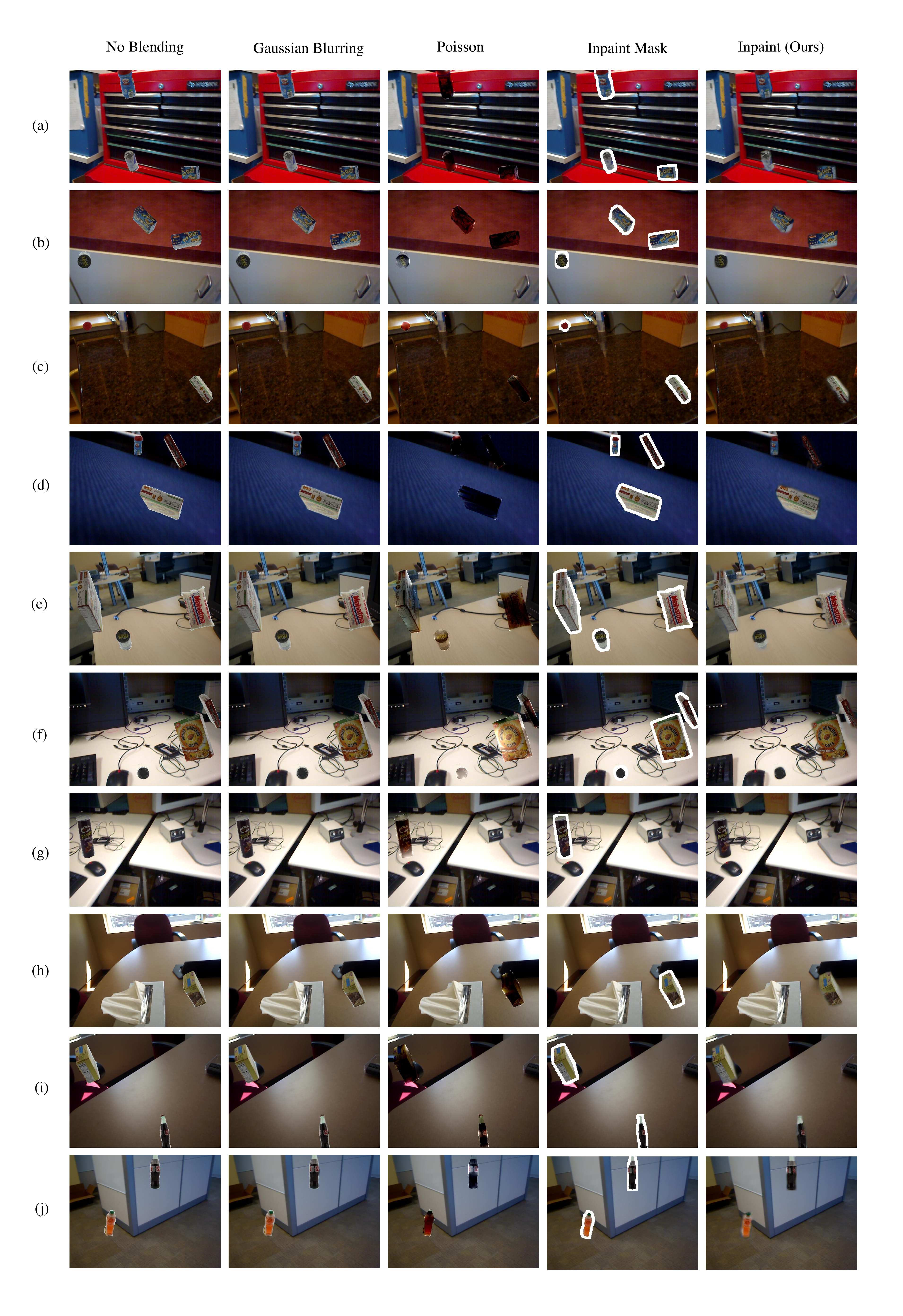}
  \caption{ Sample synthetically  generated images from the instance detection setting. Each column refers to a  different blending technique. We add the inpainting mask column to show  which pixels where filled in with the inpainting model. }
  \label{fig:instace detection}%
\end{figure*}
\subsection{Paste objects on background}
To   paste   the object on the background image we need to have an image of the object together with its corresponding mask. Our method is agnostic to how  we find the segmentation mask of the object we need to paste.
If you do not have the ground truth of the mask, you can use pre-trained foreground segmenters such as\cite{pixelobjectness} depending on the task required.
Appropriate augmentations can now be performed on the object such as rotation, scaling, illumination, etc.
Then you have to choose where to   paste   the object on the image. There are settings where a random placement\cite{dwibedi2017cut} is sufficient and there are settings where the location is  important for improving the performance of the model\cite{kassel2020using}.

The blending gap  around object mask, $\mathcal M$,  is produced in the instance segmentation case
as $Dilate(\mathcal M) \setminus Erode(\mathcal M)$ and in the foreground segmentation as $Dilate(\mathcal M) \setminus \mathcal M$ as the background of the object is correct for the scene and we do not have to worry about an incorrect background leaking into the object, Figure \ref{fig:foreground segmantation evaluation}.

 One of the  advantages of our method is that even if the object segmentation is not that accurate at the border of the object, where mistakes are prone to happen - our method can handle it.

\subsection{Fill the   blending gap  }
Once we have the background image along with the objects pasted on it with the blending gap along with the corresponding inpainting mask (Figure \ref{fig:main process} b), they can be inserted as input into the inpainting model that we trained in the first step. The inference process is fast and is performed for all images in the environment in the same way. The results may not be perfect in terms of graphics, but they do exactly the job we are interested in - cause the learning models to focus on appearance and  not recognize that it is a   pasted object.

\begin{table}[h]
    \small
    \begin{tabular}{|l|c|c|c|c|}
    \hline
    \multicolumn{5}{|c|}{\textbf{Unsupervised Foreground Segmentation methods Comparison}} \\
    \hline
\textbf{}  & \textbf{Baseline} & \textbf{Night } & \textbf{Low Frame } & \textbf{Thermal} \\
\textbf{}  & \textbf{} & \textbf{ Videos} & \textbf{Rate} & \textbf{} \\
\hline

\hline
\textbf{BSUV-Net 2.0}  & 0.962 & 0.585 & 0.790 & 0.893 \\
\textbf{SemanticBGS}   & 0.960 & 0.501 & 0.788 & 0.821 \\
\textbf{IUTIS-5 }      & 0.956 & 0.529 & 0.774 & 0.830\\
\textbf{Ours}          & \bf{0.991} & \bf{0.752} & \bf{0.893} & \bf{0.902}\\
\hline
\end{tabular}
\caption{ F-measure comparison of  state of the art unsupervised methods from CDnet dataset.}
\label{table:foreground segmantation}
\end{table}
%-------------------------------------------------------------------------

\section{Experiments}
\label{sec:intro}
Our method is general and can be applied in creating synthetic data for a wide variety of tasks.
This paper explored the capabilities of our  method in two main computer vision tasks: instance detection
and foreground segmentation.
\subsection{Training and Evaluation on Instance Detection}
To test our method in instance detection task we used the   Dwibedi et al.  \cite{dwibedi2017cut} environment as in the cut, paste, and learn paper. We use a total of
33 object instances from the BigBIRD Dataset \cite{singh2014bigbird} overlapping with the 11 instances from GMU Kitchen Dataset \cite{georgakis2016multiview}. For the  foreground/background masks of the objects, we use a pretrained foreground segmentation network \cite{pixelobjectness}. We use backgrounds from the UW Scenes Dataset \cite{lai2011large}.
We trained a inpainting model for this task on 1000 images,  600 of them were from the various background images from the UW Scenes Dataset, and another 400 randomly picked images from the BigBIRD Dataset from which all the objects that we   pasted   are. All the images were trained on inpainting irregular   holes  .
We generated a synthetic dataset with approximately 6000 images using
all the augmentations they suggested e.g.  scale, rotation, etc. where  position and the background were chosen randomly.
Each background appears roughly few times in the generated
dataset with different objects.

We created 4 different synthetic datasets that differ in terms of the pasting process e.g. No blending, Gaussian blurring, Poisson editing, and Inpainting (ours). In addition, we generated another dataset called   All Blend + same image   e.g. synthesize the  same scene with the
same object placement, and only vary the type of blending used to make the training algorithm further ignore the effects of blending as they suggested. Samples of the generated images  compared to other blending methods can be seen in Figure \ref{fig:instace detection}.  

We can see that the inpainting model  fills the missing pixels with  existing shapes of the background and combines them with the texture of the object. The border of the object is slightly blurred and the transition is smooth. This causes the detection system not to focus on the border of the object but on its appearance. In contrast, the transition between the background and the objects in the   no blending   and   gaussian blur   methods are more   sharp   and thus more noticeable. In addition, we can see that the   Poisson   blending method  often causes large parts of the object to be   occluded   in order to create a good blending.

We use a Faster R-CNN model \cite{ren2015faster} based on 
VGG-16 \cite{simonyan2014very} pre-trained weights on the MSCOCO \cite{lin2014microsoft} detection task to train the detector on the synthetic dataset we created.
For evaluation, we use the GMU Kitchen Dataset \cite{georgakis2016multiview} which contains 9 kitchen scenes with 6,728 images. We evaluate on the 11 objects present in the dataset
overlapping with the BigBIRD  \cite{singh2014bigbird} objects. We report Average Precision (AP) at IOU of
0.5 in all our experiments for the task of instance localization.
Table \ref{table:instance-detection}  shows the evaluation results.
From the results, we can see that our approach beats almost all compared blending methods  by a big margin. In addition, we also made an ablation study in the   All Blend + same image   setting. The model that was trained with data that was generated with our blending method performed better then without it.
\subsection{Training and Evaluation on Foreground Segmentation}

To test our method in foreground segmentation  we used the  setup in Kassel et al.   \cite{kassel2020using}.
The idea is to use an unsupervised segmenter to extract the foreground objects from the training frames and when  pasted it will be pasted in its
original location.
This is well suited for the static camera setting where   the objects are placed in the right location, being of the right size, color and shape, and in the right lighting conditions. 

For this purpose we use  four scene categories   from the  Change Detection 2014 (CDnet) dataset \cite{wang2014cdnet} baseline,
night videos, low frame-rate, and thermal.
The background images are pixel-wise medians of a sequence of 50 frames. The
frames used to extract objects were the same 200 used by
FgSegNet V2\cite{lim2020learning}.
For the proposal on extracting the foreground objects we
use the state of the art unsupervised method on the CDnet
dataset \cite{wang2014cdnet}, BSUV-Net v2 \cite{tezcan2021bsuv}.
For every scene, we trained a unique inpainting model that was trained on 200 training images in random masking holes and also a fixed set of   blending gap   filling that we created with the  BSUV-Net v2 results. For each scene, we generated 500 training frames. Samples from the generated frames can be seen in Figure \ref{fig:foreground segmentation}.

We can see that inpainting models fill the missing pixels  with the   right   context. e.g. shadow/light of the car, lines on the road, etc. In general, the object is inserted in a good and consistent way to blend with the background (Figure \ref{fig:main process}e).  
To test our generated data we use the  FgSegNet V2 by Lim et al. \cite{lim2020learning}, which is a state-of-the-art method in the CDnet 2014 Challenge\cite{wang2014cdnet}, SBI2015\cite{maddalena2015towards}, and
UCSD Background Subtraction \cite{UCSD}. We ignored   blending gap   pixels during the learning process since we didn't know if the inpainting model will generate foreground or background pixels. Since this approach is totally unsupervised we compared it to the state of the art unsupervised methods. Results are shown in Table \ref{table:foreground segmantation}.
 We can see that the model that trained on data that was generated with our method performs better than all the state of the art methods by a big margin,  improving the existing segmentation in the unsupervised setting.

%-------------------------------------------------------------------------
\section{Summary}
\label{sec:intro}
We presented a novel approach to    paste   objects on background images when synthesizing annotated training images for tasks like detection and segmentation. Inspired by the notion that local realism is
sufficient for training convolutions based models we focused our efforts on the connection between the   pasted   objects and the background image - we called this connection the   blending gap. Our approach uses the  inpainting  to learn the task environment and fill the   blending gap   missing pixels successfully and improves the generalization by a big margin.

%-------------------------------------------------------------------------

{\small
\bibliographystyle{ieee_fullname}
\bibliography{egbib}
}

\end{document}